\documentclass{article}

\usepackage{arxiv}

\usepackage[utf8]{inputenc} 
\usepackage[T1]{fontenc}    
\usepackage{hyperref}       
\usepackage{url}            
\usepackage{booktabs}       
\usepackage{amsfonts}       
\usepackage{nicefrac}       
\usepackage{microtype}      
\usepackage{lipsum}		
\usepackage{graphicx}
\usepackage{natbib}
\usepackage{doi}
\usepackage{array}
\usepackage{multirow}

\title{Next-Gen Traffic Surveillance: AI-assisted Mobile Traffic Violation Detection System}

\author{
\begin{minipage}[t]{0.3\textwidth}
  \centering
  \href{https://orcid.org/0009-0001-8326-1710}
  {\includegraphics[scale=0.06]{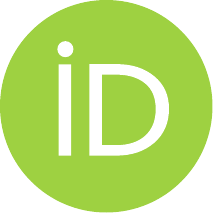}\hspace{1mm}Dila ~Dede}\\
  Department of Electronics and Communications Engineering\\
  Istanbul Technical University\\
  Maslak, Istanbul, 34485, Turkey \\
  \texttt{diladede2006@hotmail.com}
\end{minipage}
\begin{minipage}[t]{0.3\textwidth}
  \centering
  \href{https://orcid.org/0009-0000-0991-0578}
  {\includegraphics[scale=0.06]{orcid.pdf}\hspace{1mm}Mehmet Ali~Sarsıl}\\
  Department of Electronics and Communications Engineering \\
  Istanbul Technical University \\
  Maslak, Istanbul, 34485, Turkey \\
  \texttt{sarsil18@itu.edu.tr}
\end{minipage}
\begin{minipage}[t]{0.3\textwidth}
  \centering
  {\hspace{1mm}Ata~Shaker} \\
  Department of Electronics and Communications Engineering\\
  Istanbul Technical University\\
  Maslak, Istanbul, 34485, Turkey \\
  \texttt{shaker20@itu.edu.tr}
\end{minipage}\\ \\
\begin{minipage}[t]{0.3\textwidth}
  \centering
  {\hspace{1mm}Olgu~Altıntaş} \\
  Department of Electronics and Communications Engineering\\
  Istanbul Technical University\\
  Maslak, Istanbul, 34485, Turkey \\
  \texttt{altintasol19@itu.edu.tr}
\end{minipage}
\begin{minipage}[t]{0.3\textwidth}
  \centering
  \href{https://orcid.org/0000-0001-7226-4898}{\includegraphics[scale=0.06]{orcid.pdf}\hspace{1mm}Onur~Ergen}\thanks{Corresponding author: Onur Ergen, Email: oergen@itu.edu.tr.} \\
  Department of Electronics and Communications Engineering\\
  Istanbul Technical University\\
  Maslak, Istanbul, 34485, Turkey \\
  \texttt{oergen@itu.edu.tr}
\end{minipage}
}

\date{}

\renewcommand{\headeright}{}
\renewcommand{\undertitle}{Research Paper}
\renewcommand{\shorttitle}{}

\hypersetup{
pdftitle={A template for the arxiv style},
pdfsubject={q-bio.NC, q-bio.QM},
pdfauthor={David S.~Hippocampus, Elias D.~Striatum},
pdfkeywords={First keyword, Second keyword, More},
}

\begin{document}
\maketitle

\begin{abstract}
Road traffic accidents pose a significant global public health concern, leading to injuries, fatalities, and vehicle damage. Approximately 1,3 million people lose their lives daily due to traffic accidents \citep{who2022roadtrafficinjuries}. Addressing this issue requires accurate traffic law violation detection systems to ensure adherence to regulations. The integration of Artificial Intelligence algorithms, leveraging machine learning and computer vision, has facilitated the development of precise traffic rule enforcement. This paper illustrates how computer vision and machine learning enable the creation of robust algorithms for detecting various traffic violations. Our model, capable of identifying six common traffic infractions, detects red light violations, illegal use of breakdown lanes, violations of vehicle following distance, breaches of marked crosswalk laws, illegal parking, and parking on marked crosswalks. Utilizing online traffic footage and a selfmounted ondash camera, we apply the YOLOv5 algorithm's detection module to identify traffic agents such as cars, pedestrians, and traffic signs, and the strongSORT algorithm for continuous interframe tracking. Six discrete algorithms analyze agents' behavior and trajectory to detect violations. Subsequently, an Identification Module extracts vehicle ID information, such as the license plate, to generate violation notices sent to relevant authorities.
\end{abstract}


\section{Introduction}
Road traffic accidents are the main cause of death for young people and the eighth major cause of all deaths globally with a predestined 1.24 million deaths every year \citep{mohammed2019review}. Many governments throughout the world rely on universal traffic laws and enforcement agencies to mitigate the effects of driver behavior and to enhance road safety \citealp{bates2012effectiveness}. Traffic law enforcement agencies rely on two main related functions of traffic law enforcement, which are apprehension and deterrence. After having established traffic laws, government bodies must police these laws to apprehend offenders, who later face penalties and sanctions to deter them from committing illegal activity in the future. This road safety system relies on accurate and universal traffic law violation detection methods. 
The main objective of implementing traffic laws is thus to enable vehicles to move safely and attentively on existing roads. By applying suitable traffic regulations and control measures, as well as having roads that are well-equipped with the appropriate traffic management facilities, driver behavior can be monitored, and reckless driving can be sanctioned.

Typically, single-entity frameworks such as the traffic police are used to monitor driver anomalies and dangerous driving behavior. This in-person detection method is inefficient and cannot patrol a large percentage of roads. Intelligent Transportation Systems (ITS) \citep{qureshi2013survey} are being adopted by many cities to improve coordination, adaptivity, and automated response for traffic transportation policy optimization, thus increasing “smartness” and efficiency within cities \citep{europeancommission}. Many ITS’ applications are focused on improvements in the existing traffic detection infrastructure, notably mobile detection for each traffic agent on the road. Therefore, Intelligent Transportation Systems have played a major role recently in reducing risks, high accident rates, and traffic congestion and on the other hand increasing safety, reliability, travel speeds, traffic flow, and satisfied travelers \citep{alam2016introduction, lin2017intelligent}.
With the rise of ITS and other technological advancements, new intelligent vehicle violation recognition methods have been proposed. These new methods rely predominantly on the use of Artificial Intelligence, specifically Machine Learning and computer vision algorithms. The in-tune combination of machine learning and computer vision techniques to form algorithms for resolving this problem and enforcing traffic laws has shown promise. Models using CNN, infrared imaging analysis using FNN \citep{pavlidis2000vehicle}, and IVIoT \citep{li2015road} have been able to detect vehicles on the road, notably using frame-by-frame comparisons. The scope of these models has been advanced to include the detection of traffic law violations. Models such as ‘Lane Quest’ developed by M. Fuad et al. \citep{arnob2020intelligent} are specialized in the task of detecting a vehicle’s lane and the surrounding vehicles in the given frame. Aaron Christian et al. presented a machine vision algorithm to detect traffic violations specifically swerving and blocking the pedestrian lane \citep{christian2016machine}. These AI-based models aim to detect singular traffic violations and rely globally on external traffic camera footage. In this paper, we present a mobile traffic violation detection model that incorporates various computer vision modules to detect 6 separate traffic violations, using self-acquired on-dash camera footage as well as internet footage. We developed six distinct algorithms for each infraction by looking at agent behaviors and their interactions with traffic signs. When violations are detected, an Identification Module is activated, which gathers agent data and notifies the appropriate authorities of the violations, helping to improve traffic safety and rule enforcement.
\begin{figure}
	\centering
        \includegraphics[width=14cm]{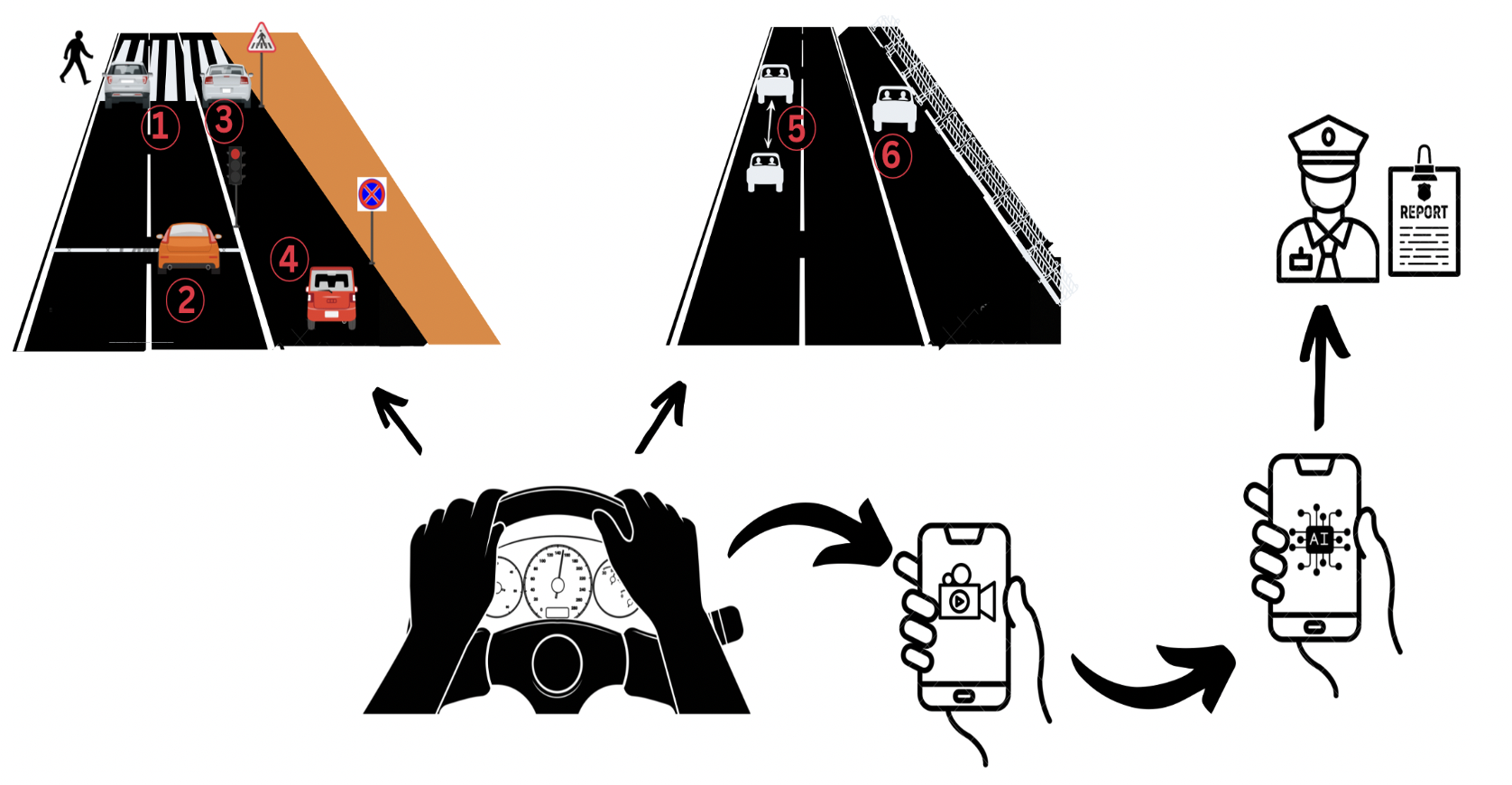}
	\caption{Potential application of the proposed method to enable democratized mobile traffic violation detection and notification system using a smartphone. Pedestrians, drivers, or any kind of traffic participants can take footage involving one of 6 traffic violations committed by the on-road agents, such as cars, buses, trucks, motorcycles, etc. The camera footage taken by the participant can be uploaded to the system. Then, following the verification of the violation using the fine-tuned algorithm, the identity of the agent is sent to the relevant authorities.}
	\label{fig:fig1.png}
\end{figure}
This combinatorial model can prove efficient in identifying traffic law violations using footage taken on a mobile phone. This footage can be recorded by any vehicle or pedestrian on the road and then analyzed by the proposed algorithm to detect and confirm certain traffic law violations. The visibility of the footage input into the algorithm is a prerequisite to deducing a violation. This rapid violation detection will allow drivers to notice and report reckless driving using their mobile phones at any time or place on the road. This algorithm can be used in concordance with government authorities to allow for the democratization of complaint systems, thereby improving the apprehension and deterrence of traffic law violators.

\section{Methodology}
\label{sec:headings}
\subsection{Data Collection}
The YouTube platform \citep{youtube} is the primary source of collected videos involving scenes of traffic violations from an on-dash perspective. Nevertheless, the data was not confined solely to videos from sources in Türkiye but also spanned to countries in Asia, Europe, and America. Therefore, various other videos from diverse countries were incorporated to develop the algorithm.
A self-acquired on-dash mounted camera was also used by our team to acquire independent footage to expand our collection of sample videos. The camera, GoPro Hero Camera 5GPR/CHDHA-301-EU, was mounted on the dashboard of a personal car to facilitate the collection of continuous video recordings while driving in various districts in Istanbul. The total duration of the collected videos is approximately 10 hours. 

\subsection{Machine Learning}
The detection of traffic agents, mainly motorized vehicles, in a single frame, is the initial paramount step to ensure the position information of in-frame vehicles. After having detected each individual agent within a singular frame, the algorithm can then analyze a video, composed of a succession of frames to identify the nature of a potential violation.
\subsubsection{Object Detection of Traffic Agents using YOLOv5}
Object detection systems are typically composed of three primary components: the Backbone, the Neck, and the Head. The Backbone component is responsible for extracting essential features from input images \citep{diwan2023object}. Popular choices for backbones include VGG \citep{simonyan2014very}, ResNet \citep{he2016deep}, DenseNet \citep{huang2017densely}, and CSPDarknet53 \citep{wang2020cspnet}, all recognized for their robust feature extraction capabilities in tasks like object detection and classification. The Neck component is positioned between the Backbone and the Head and serves as a feature extraction enhancer. It plays a pivotal role in the detection network, often involving multiple bottom-up and top-down pathways. Various path-aggregation techniques, such as FPN \citep{lin2017feature}, BiFPN \citep{le2020efficientdet}, and ASFF \citep{jiang2020fassd}, are used to improve feature fusion by employing operations like upsampling, downsampling, splicing, and dot product. The Head component utilizes the features obtained from the Backbone to perform target localization and classification. Object detection methods are categorized into one-stage and two-stage detectors. YOLO (You Only Look Once) is a well-known example of a one-stage detector that simultaneously predicts bounding boxes and target classes, offering speed advantages but potentially lower detection accuracy compared to two-stage detectors. In the context of YOLOv5, it primarily employs the CSPDarkNet53 structure, utilizing components such as the Convolutional (Conv) layer, C3 layer, and SPPF layer in the backbone network. The Conv layer includes convolution, batch normalization, and the SiLU function. The C3 module efficiently reduces model parameters and enhances inference speed by using residual connections. SPPF, an improved version of the original SPP module, replaces max-pooling layers of various sizes with three max-pooling layers of size 5 × 5, which aids in feature fusion and operational speed. YOLOv5's Neck component incorporates the Path Aggregation Network (PANet), an extension of the Feature Pyramid Network (FPN), to enhance localization across different scales by introducing bottom-up pathways that transmit positional information from lower to deeper levels after top-down feature fusion in FPN.

YOLOv5 encompasses five derived models, denoted as YOLOv5n, YOLOv5s, YOLOv5m, YOLOv5l, and YOLOv5x. While these models maintain a consistent architecture, they differ in terms of their dimensions (Zhang, 2023). For a thorough depiction of the overall structure of the initial YOLOv5 model, refer to Figure 2.    

\begin{figure}
	\centering
        \includegraphics[width=12cm]{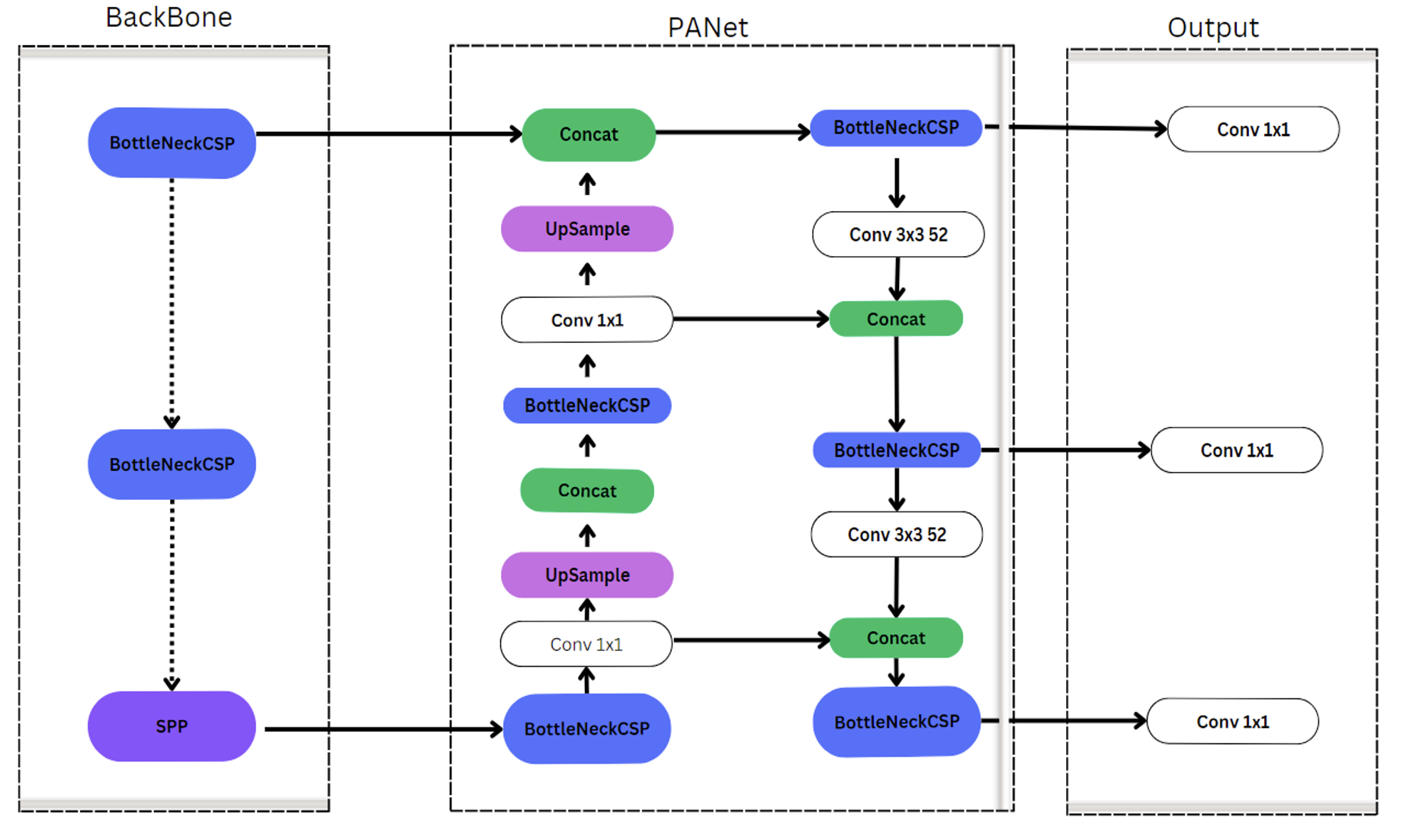}
	\caption{An overview of YOLOv5 framework.}
	\label{fig:fig1.png}
\end{figure}

YOLOv5 is used to carry out the detection of motor vehicles in frames. YOLOv5 is the fifth version of the deep learning frameworks of the You Only Look Once (YOLO) family of computer vision model developed by Ultralytics. Built on the PyTorch framework, YOLOv5 has garnered immense popularity for its versatility, ease of use, and high performance in various applications, namely object detection, instance segmentation, and image classification.

A variety of sets of pre-trained weights for the YOLOv5 model are available online. As an object detection module, the weights of the model “yolov5su.pt” with 9.1 million parameters trained on the COCO dataset were used \citep{ultralytics2023guide}. The COCO dataset is a remarkable resource in computer vision, encompassing 200,000+ images and annotations of objects from 80+ object categories. COCO serves as a benchmark for evaluating cutting-edge computer vision algorithms and models \citep{lin2014microsoft}. The classes to be detected by the YOLOv5 module can be input manually by the user. Among the available 80 classes, the module trained on the COCO dataset can detect 4 classes of interest which are the selected agents on which the model performs its detection. These 4 agents are cars, motorcycles, buses, and trucks. Furthermore, the YOLOv5 module is also capable of detecting 3 traffic law infractions: Illegal Passing on Traffic Lights, Illegal use of Breakdown lanes, and Violation of Following Distance.

To further detect an additional three violations, the model required supplemental datasets to identify two specific traffic signs, namely the "no stopping" sign and the pedestrian crossing sign. Identifying these two traffic signs would allow the model to be able to detect three more traffic violations: Pedestrian Crossing Violation, Illegal Stopping Violation, and Marked-Crosswalk Parking Violation. Therefore, the YOLOv5 model underwent retraining to allow for the model to expand its traffic law infraction detection range from three to six. This retraining process involved the creation of an expanded dataset containing new images of these signs. To accomplish this, online image databases containing these two traffic signs, complete with annotations and data labeling, were assembled. Traffic scene images featuring these signs were sourced from various databases, including the Chinese Traffic Sign Database \citep{chinesetrafficsigndb}, the BelgiumTS Dataset \citep{timofte2011multiview}, the Turkish Traffic Sign Dataset (TTSD) \citep{ttsd}, and the Russian traffic sign images dataset (RTSD) \citep{popov2020data} databases. The final dataset incorporated a total of 12,977 images from each previously mentioned dataset to showcase these signs, consisting of 6,514 "no stopping" signs and 6,463 crosswalk signs. Additionally, 1,226 random images were selected from these databases to serve as background images.  These 1,226 background images were incorporated during training to help the model learn about scenes that do not contain the desired objects, ultimately improving the model's ability to reduce false positive predictions \citep{ultralytics2023tips}. The dataset was divided into a training set (90\%) and a validation set (10\%), resulting in 11,656 images for training and 1,321 for validation. The YOLOv5 architecture was trained with these specifications. The same architecture as "yolov5su.pt" was employed, with initial image resizing to 415, a batch size of 16, and a total of 30 epochs. 

\subsubsection{Across-frame Object Tracking}

The SORT algorithm addresses multi-object tracking by incorporating both historical and current video frames. It tackles the challenge of predicting object motion trajectories and correlating them with video frame data using the Kalman Filter and the Hungarian algorithm \citep{leal2015motchallenge}. The Kalman filter \citep{kalman1960new} is a recursive mathematical tool used for tracking in SORT. Each track in SORT is associated with a Kalman filter that predicts the future state of the object being tracked. The Kalman filter combines the object's current state, including its position and velocity, with the incoming measurements to estimate the new state of the object more accurately. This process of prediction and correction is crucial for maintaining the track's accuracy and handling noise or missing measurements. The Hungarian algorithm \citep{priya2019hungarian} is used in SORT for solving the data association problem, also known as the assignment problem. The goal is to assign detected objects to existing tracks or create new tracks for unassigned objects while optimizing the cost function. In SORT, the Hungarian algorithm is employed to find the best assignment of objects to tracks, considering the distance or similarity between object detections and track predictions. It provides an efficient and optimal solution for this combinatorial problem, ensuring that the correct associations are made while minimizing the overall cost. This approach aims to achieve frame-to-frame correlation in object detection. However, issues arise when objects become occluded, leading to a mismatch between the predicted trajectory by the Kalman Filter and the results detected by the object detector. This can result in hindered object tracking and a significant number of target ID switches. DeepSORT, an extension described in \citep{wojke2017simple}, enhances SORT by introducing a pre-trained Convolutional Neural Network (CNN) to preserve appearance features from the last 100 frames of each trajectory. This innovation effectively mitigates ID switching caused by occlusions. Additionally, DeepSORT incorporates cascade matching and novel trajectory confirmation techniques to optimize the alignment between the predicted trajectory and the object in the current frame. In \citep{yunhao2022strongsort}, Du, Y. et al. introduce strongSORT, which further builds upon these advancements. They propose two lightweight plug-and-play algorithms: AFLink and GSI. The AFLink model focuses on associating short trajectories with complete trajectories using a fully connected model that does not rely on appearance information. On the other hand, GSI tackles the issue of missing object detections by simulating nonlinear motion and achieving more accurate object positioning through Gaussian regression. Importantly, it retains motion information during the regression process, ensuring a more robust tracking system. In our paper, we leverage the strongSORT algorithm as a crucial component in our approach for cross-frame object tracking, complementing the capabilities of the YOLOv5 object detection model. The online available weights of the pre-trained CNN required for the incorporation of strongSORT with YOLOv5 were utilized \citep{yunhao2022strongsort}.
\subsubsection{License Plate Detection Module - WPOD-NET}

License plates (LPs) are inherently flat and rectangular objects that are adhered to vehicles for identification purposes. The detection of LPs on vehicles is achieved through the utilization of the Warped Planar Object Network (WPOD-NET). This network has been trained to recognize license plates from Europe, the USA, and Brazil, but it can be adapted to identify plates from other regions as needed. Despite being significantly slower (20 times) than OpenCV, WPOD-NET can still process images in real-time, offering increased resilience and exceptional performance \citep{hung2020improving}. When presented with an image of a vehicle, the Warped Planar Object Detection Network (WPOD-NET) actively seeks out license plates and calculates an affine transformation for each detection. This transformation rectifies the distorted license plate into a rectangular shape resembling the front view of the plate. This process is initiated by employing the vehicle detection module YOLOv5 to frame the vehicle image within its bounding box. Subsequently, WPOD-NET, a Convolutional Neural Network-based system, is engaged to pinpoint the license plate. This network is trained to identify license plates despite various distortions and computes coefficients for an affine transformation that corrects the distorted license plate, rendering it in a rectangular shape. In its initial stages, the network is fed with the resized output from the vehicle detection module. The forward pass yields an 8-channel feature map that encodes probabilities for object presence and parameters for affine transformations. To extract the distorted LP, a hypothetical square of fixed dimensions is considered, centered within a cell (m, n) in the image. If the object probability for this cell exceeds a predefined detection threshold, a portion of the regressed parameters is employed to construct an affine matrix that transforms the hypothetical square into a recognizable LP area. This process simplifies the task of unwarping the LP, aligning it both horizontally and vertically. In its initial stages, the network is fed with the resized output from the vehicle detection module. The forward pass yields an 8-channel feature map that encodes probabilities for object presence and parameters for affine transformations. To extract the distorted LP, a hypothetical square of fixed dimensions is considered, centered within a cell (m, n) in the image. If the object probability for this cell exceeds a predefined detection threshold, a portion of the regressed parameters is employed to construct an affine matrix that transforms the hypothetical square into a recognizable LP area. This process simplifies the task of unwarping the LP, aligning it both horizontally and vertically.

\subsubsection{License Plate Optical Character Recognition - Image Processing and MOBILENET}

An algorithm with multi-step image processing was employed based on the algorithm proposed by Silva et al. for the extraction and identification of characters from a detected license plate image by WPOD-NET \citep{silva2018license}. The license plate image is enhanced through grayscale conversion, Gaussian blur, and thresholding. These morphological operations can further refine character regions. Contours of individual characters are detected and sorted from left to right. Each character region is then isolated, resized to a predefined standard (80x80 pixels), and thresholded for clarity. The characters are recognized using a MobileNet-based model with shared pre-trained weights, and the predicted characters are aggregated into a final character string representing the license plate. MobileNets \citep{howard2017mobilenets} are efficient models designed for mobile and embedded vision applications, offering a streamlined architecture with depth-wise separable convolutions. They feature adaptable hyper-parameters, striking a balance between speed and accuracy, making them ideal for various applications. Notably, MobileNets exhibit strong performance in tasks like ImageNet classification and are effective for tasks such as license plate optical character recognition \citep{ashrafee2022realtime}.    

The comprehensive approach for license plate character recognition, as can be seen in Figure 3, offers a valuable tool for various applications, ensuring accurate and efficient identification of license plate characters with the benefit of leveraging pre-trained model weights.  

\begin{figure}
	\centering
        \includegraphics[width=14cm]{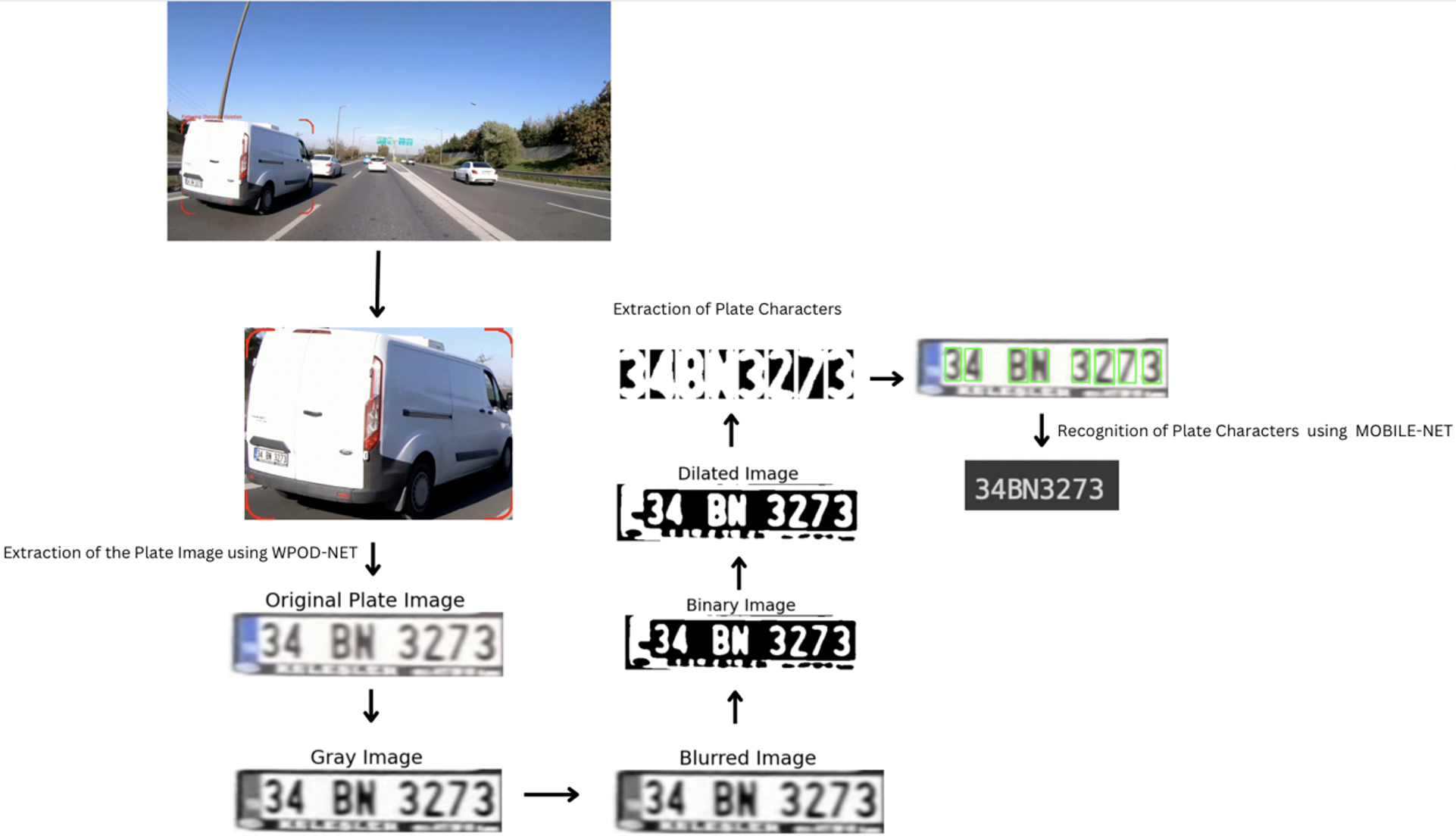}
	\caption{The algorithm followed for vehicle identification: WPOD-NET license plate extraction, a multi-step image processing operation, and MobileNet license plate character recognition.}
	\label{fig:fig1.png}
\end{figure}

\subsection{Traffic Violation Detection Algorithms}

\subsubsection{Illegally Running a Red Light}
To ensure orderly and lawful passage of vehicles at crossroads, a red, yellow, and green color system is used for traffic lights \citep{retting1998red, retting1999prevalence}. The most frequent and major traffic law infraction consists of running a red light. Detecting this violation relies on a model capable of identifying and tracking traffic lights and vehicles. In this study, the YOLOv5 model is employed to extract relevant objects from a mobile video.  The devised algorithm must first determine the inclination of the traffic light, which corresponds to its classification as a ‘vertical’ or ‘horizontal’ traffic light. If the bounding box corresponding to the traffic light is longer in the vertical direction (y-axis), the traffic light must be vertically aligned along the x-axis. Conversely, if the traffic light’s bounding box is wider in the horizontal direction, then the traffic light must be horizontally aligned along the x-axis. Based on this morphological information, a splitting algorithm splits the bounding box of the traffic light either vertically or horizontally into three equal pieces, corresponding to the three possible colors of the traffic light: Red, Yellow, and Green. To determine the traffic light’s color information, successive image processing steps must be applied to the image of the traffic light. All splits contain Red(R), Green(G), and Blue(B) channels. The pixel intensities of each channel (RGB) are divided by the pixel intensity of the dominant channel. For instance, the pixel intensities of the red and blue channels in the split that correspond to a dominant green light are divided by the pixel intensity of the green channel, whereas the pixel intensities of the green and blue channels in the split that corresponds to a dominant red light are divided by the pixel intensity of the red channel.

Then, the obtained splits are converted to grayscale images to perform the subsequent image processing steps. At this stage, if the traffic light is red, the pixels that constitute the red-light split are ‘brighter’ compared to the other two splits. Similarly, when the traffic light is green, the green light split appears brighter in the grayscale channel. The grayscale image is then thresholded by a determined pixel intensity value of 40. The split images are converted to binary (black or white) images. A dominant white pixel value corresponds to a state of illumination, which implies that the traffic light is indeed on. Inversely, a dominant black pixel value implies that the other two lights are off.  By comparing the number of white pixels in each split, its state of illumination can be confirmed. This conversion allows the algorithm to further confirm the dominant pixel value of the traffic light in suboptimal lighting conditions by intensifying the pixel values. This determination process can be error-prone in a single frame due to lighting and perspectives. Therefore, the decision regarding the color of the light is made by examining the determined color in the subsequent 10 frames. The grayscale conversion and binary thresholding processing steps allow the algorithm to filter out unnecessary color information to confirm the state of illumination and the color of the traffic light. This method ensures accurate identification of the traffic light's color, aiding in the detection of red-light violations by vehicles. Finally, the algorithm focalizes vehicles with a decreasing bounding box size in the subsequent frames of the video, which indicates that the vehicle is accelerating or continuing at the same pace relative to the traffic light and that it has therefore passed the light. The algorithm can then determine the legality of the vehicle’s passage. In the case of an illegal crossing, the vehicle’s plate is extracted using the license plate detection and recognition algorithm, and the characters on the plate are identified and reported to the authorities.

\begin{figure}
	\centering
        \includegraphics[width=14cm]{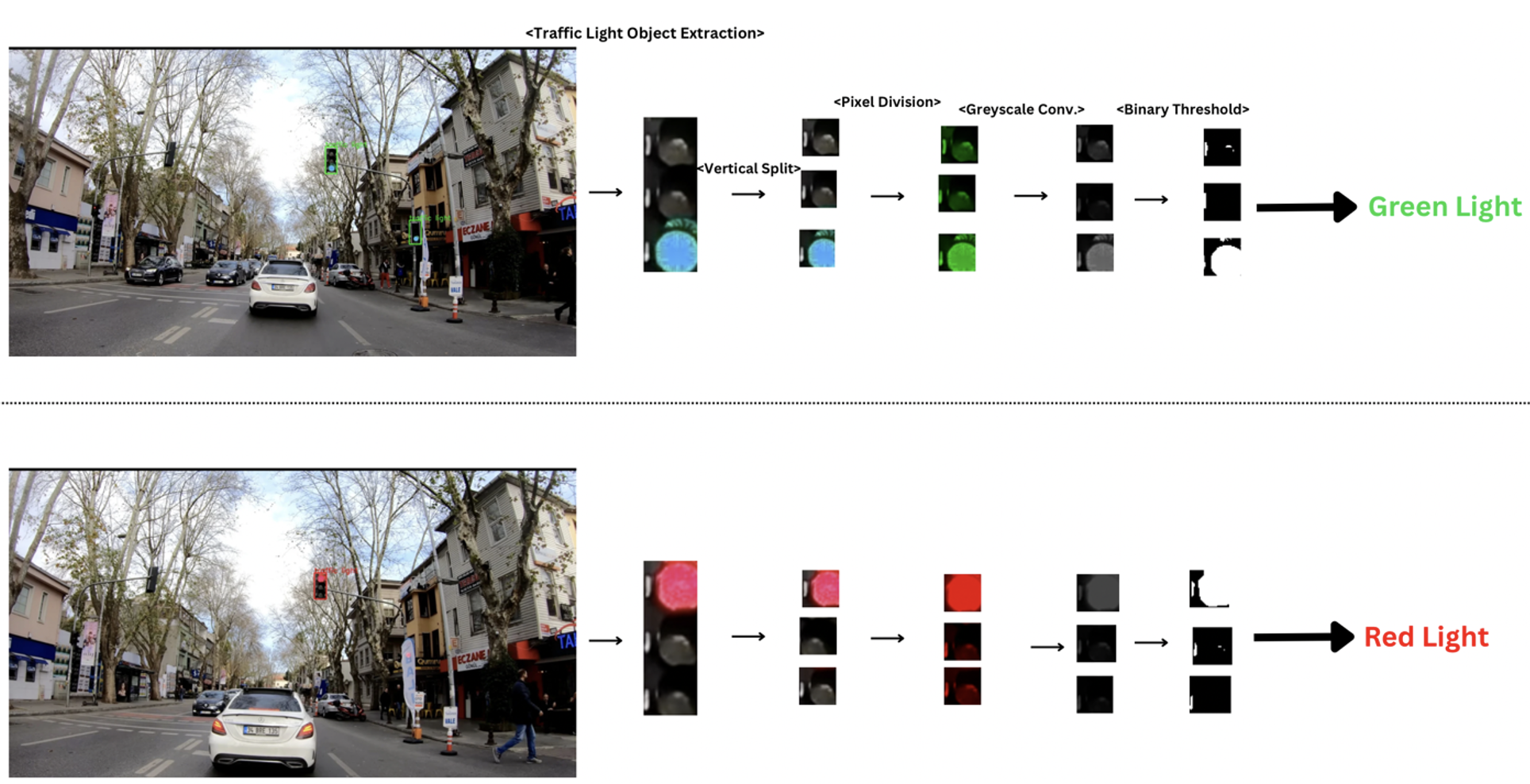}
	\caption{Determination of state of illumination of a traffic light: YOLOv5 carries out the detection of the traffic light object, a consecutive image processing steps are incorporated to derive the color of the traffic light.}
	\label{fig:fig1.png}
\end{figure}
\subsubsection{Illegal Use of Breakdown Lanes}

Breakdown lanes are designated for vehicles with emergencies or mechanical problems. Their use outside of an emergency circumstance ensues an illegal use of breakdown lanes \citep{feng2018research}. Lane detection algorithms with pure image processing techniques are successful in traffic environments with favorable lighting conditions \citep{ke2023combining}. These algorithms utilize the location of vehicles relative to their lane using their bounding box coordinates. Conversely, these methods fail in traffic environments with poor visibility of road markings, which represent most cases. In this scenario, the model divides the total frame into three vertical frames. The across-frame change of the bounding box area of each vehicle in the entire scene is examined to determine each vehicle’s speed. If the speeds of the vehicles in the second and third splits are higher than the average of the speeds of all vehicles in the scene, the vehicles situated in the second and third splits must have an abnormally high speed compared to the average speed of other vehicles in the whole frame. Therefore, the vehicles in these two splits are regarded to have used breakdown lanes due to their significantly higher speed relative to other slower vehicles. This method works efficiently in cases where the vehicles in the entire frame are stuck in traffic and are therefore moving slowly, whereas the vehicle using a breakdown lane has been able to accelerate in speed. In the case of an illegal use of breakdown lanes, the vehicle’s plate is extracted using the identification module, and the characters on the plate are reported to the authorities.
\subsubsection{Violation of Following Distance}
To ensure the reduced risk of collision in the event of an accident, the minimum distance between vehicles in the same lane must be above a certain threshold based on the speed limit designated for the road. This concept is known as the 3-second rule and is enforced to give other drivers on the road adequate time to react in the case of an incident \citep{mcdonald2023following}. The 3-second rule is described as how far you can travel within three seconds at a set speed. Keeping this distance between cars is the safest way to avoid being involved in a rear-end collision. 

Violation of the following distance rule can be detected using a basic analytical approach. A camera can observe vehicles merging from the right into left lanes or vice versa. When a vehicle appears from one of the two sides of the frame, the algorithm's time module is activated.  If the detected vehicle arrives at the position of the vehicle that was initially in front of it in under 3 seconds, this indicates that the vehicle accelerated and therefore violated the 3-second following distance rule. Moreover, the algorithm operates on the base assumption that during the entirety of the footage, the vehicle with the mounted camera on the dash is traveling at a constant speed that conforms to the speed limit of the road. Single-frame vehicle detection and across-frame tracking are implemented with the YOLOv5 and strongSORT modules, respectively. In the case of a violation of the following distance rule, the vehicle’s plate is extracted using the vehicle identification module, and the characters on the plate are reported to the authorities.

\subsubsection{Pedestrian Crossing Violation}
Pedestrians intending to cross the street using marked crosswalks have priority over ordinary vehicles on the road \citep{ragland2007driver}. In normal circumstances, drivers must stop before crosswalks and allow pedestrians to cross to the other side \citep{hamed2001analysis, koh2014safety}.

To detect this violation, the model first needs to detect pedestrian crossing signs and pedestrians in proximity to these signs. Although YOLOv5 is used to detect vehicles and pedestrians, the model must be retrained on a dataset composed of pedestrian crossing signs to incorporate them into the list of detectable object classes. 

After having detected these traffic signs, the algorithm then assesses whether a pedestrian is present near the crosswalk sign. The YOLOv5 detects a pedestrian which is followed by the analysis of the change in the x-axis location of the center of the pedestrian’s bounding box within the frame. The illegal passing of a vehicle at a pedestrian traffic sign is determined by assessing whether the vehicle stopped at the pedestrian sign. This violation can be detected using the speed of the vehicle once it approaches the pedestrian crossing sign. The algorithm refers to the change in the bounding box of the pedestrian to assess the speed of the car. The algorithm is therefore able to detect pedestrians waiting to cross the street and analyze the speed of the vehicle once it passes the pedestrian crosswalk sign. In the case of a pedestrian crossing violation, the vehicle’s plate is extracted, and the characters on the plate are identified and reported to the authorities.

\subsubsection{Illegal Parking Violation}
Illegitimate parking alongside roads is an extremely prevalent violation observed on a global scale \citep{chowdhury2018automated, morillo2014street}. These signs are predominately situated in urban environments with a lack of parking spaces. The YOLOv5 model was trained on a dataset formed by combining various online datasets of “no stopping or parking” traffic signs as aforementioned. When this traffic sign is detected within one of the two vertical frames of the footage, the algorithm reduces its detection range to the frame in which the traffic was detected. This allows the YOLOv5 model to avoid false positives and to detect illegal parking spots. The algorithm examines the temporal changes in the positioning of vehicles with respect to the identified traffic signs as well as the filming vehicle. In the case of an illegal stopping violation, the vehicle’s plate is extracted, and characters on the plate are identified and reported to the authorities. In the case of an illegal stopping violation, the vehicle’s plate is extracted, and characters on the plate are identified and reported to the authorities.

\subsubsection{Marked-Crosswalk Parking Violation}

Parking on a marked crosswalk is a prevalent violation frequently observed in traffic scenarios \citep{polak1992illegal, zoika2021causal}. Such violations not only hinder the passage of pedestrians using marked crosswalks but also obstruct the visibility of marked crosswalks due to the vehicles illegal parking location, making accidents more likely. Once the algorithm has detected a marked crosswalk sign, it becomes sensitive to vehicles on the corresponding side of the sign. The YOLOv5 model was trained with an extensive data set composed of traffic sign images and can detect traffic signs indicating a marked crosswalk. Temporal changes in the positioning of vehicles with respect to the identified traffic signs and to the vehicle where the camera was mounted are utilized to conclude the detection of the violation.

In the case of a marked crosswalk parking violation, the vehicle’s plate is extracted, and characters on the plate are identified and reported to the authorities. In the case of a marked crosswalk parking violation, the vehicle’s plate is extracted, and characters on the plate are identified and reported to the authorities.

\section{Results and Discussion}

This study is comprised of a set of 6 algorithms capable of detecting six distinct traffic violations of vehicles on the road. They use YOLOv5 as a common one-frame object detection module, strongSORT as an across-frame object tracker, WPOD-NET as a license plate extractor, and a set of image processing steps that segment the characters on the detected plate which then are then identified individually using the MobileNet character recognition model. These six algorithms were tested on a series of videos taken from the YouTube platform and personal footage. The figures in the supplementary figures section illustrate one example for each type of violation detection.

The six algorithms were able to detect 23 vehicles that violated the rules in 14 scenes, which indicates 100\% accurate detections by the algorithms. A video made up of the footage was designed and is available on the YouTube platform: \href{https://www.youtube.com/watch?v=VvTLdwCU-cI}{\textbf{Mobile Traffic Violation Detection System}}.
\begin{table*}[h]
    \centering
    \caption{Dataset details for evaluating the performance of the developed algorithms.}
    
    \begin{tabular}{|c|c|c|} 
       \hline
    Type of traffic violations &  Number of Scenes &Detected Vehicles \\
    \hline 
    
    Illegally Running a Red Light & 4 & 4 \\
    \hline
    Illegal Use of Breakdown Lanes & 3 & 6  \\
    \hline 
    Pedestrian Crossing Violation
     & 1 & 1  \\
    \hline
    Illegal Parking Violation 
     & 2 & 6 \\
    \hline 
    Violation of Following Distance& 2 & 2 \\
    \hline Marked-Crosswalk Parking Violation & 2 & 4 \\
    \hline
    Total & 14 & 23 \\
        \hline
    \end{tabular}
    \label{tab:yourlabel}
\end{table*}

\section{Conclusion}
In summary, this paper introduces a mobile model designed to detect six different traffic violations, with the added advantage of enhancing the efficiency and practicality of conventional detection models. This innovative model holds potential for adoption by both traffic surveillance agencies and individual drivers, enabling them to identify and report instances of reckless driving. Beyond complementing existing traffic monitoring tools, the methods detailed in this paper represent an initial stride towards establishing a comprehensive reporting and notification system that empowers citizens to connect directly with the relevant authorities, free from any hindrances.

The foundation laid here for traffic violation detection can be extended to create a universally applicable model, incorporating even larger datasets of video footage. This advanced model would enable the real-time tracking of the behavior of each visible entity within the traffic scene and promptly identify any violations.
Furthermore, we envision promising future applications, including the development of a comprehensive complaint system that can be accessed by all traffic personnel equipped with dash cameras. This system would provide traffic agents with the means to detect and report traffic infractions systematically, contributing to a safer road environment within cities.

\bibliographystyle{unsrtnat}
\bibliography{references}  

\section{Supplementary Figures}
\begin{figure}[b]
	\centering
    \includegraphics[width=14cm]{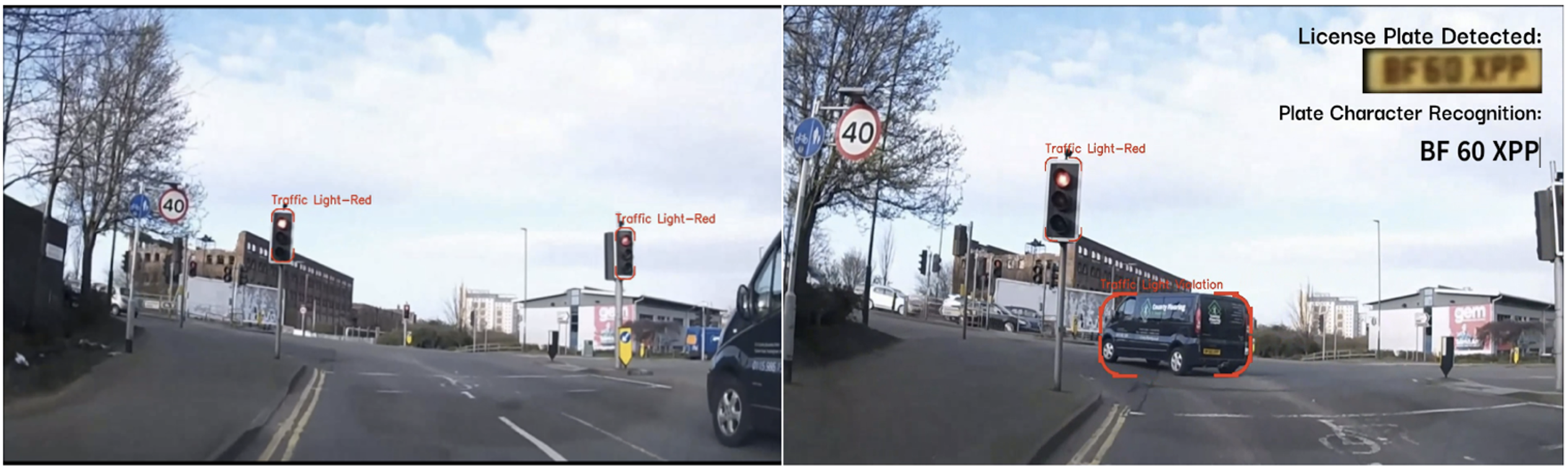}
	\caption{Detection of Illegally Running a Red Light}
	\label{fig:fig1.png}
\end{figure}

\begin{figure}
	\centering
        \includegraphics[width=14cm]{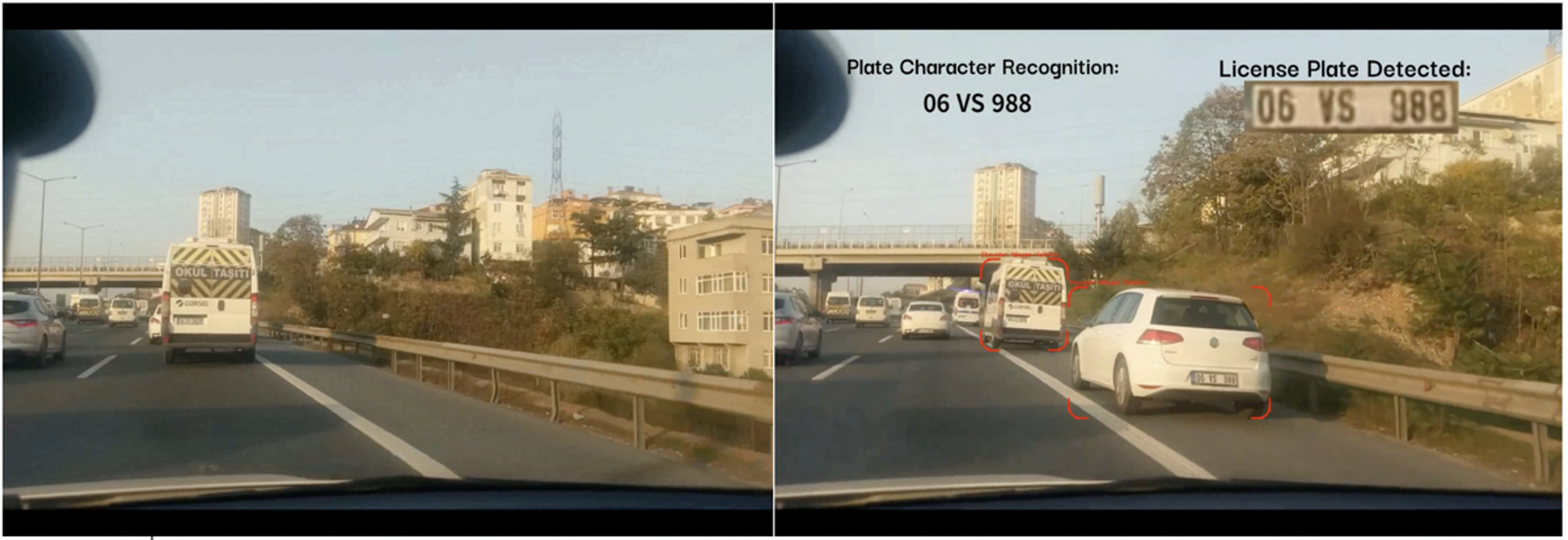}
	\caption{Detection of Illegal Use of Breakdown Lanes}
	\label{fig:fig1.png}
\end{figure}

\begin{figure}
	\centering
        \includegraphics[width=14cm]{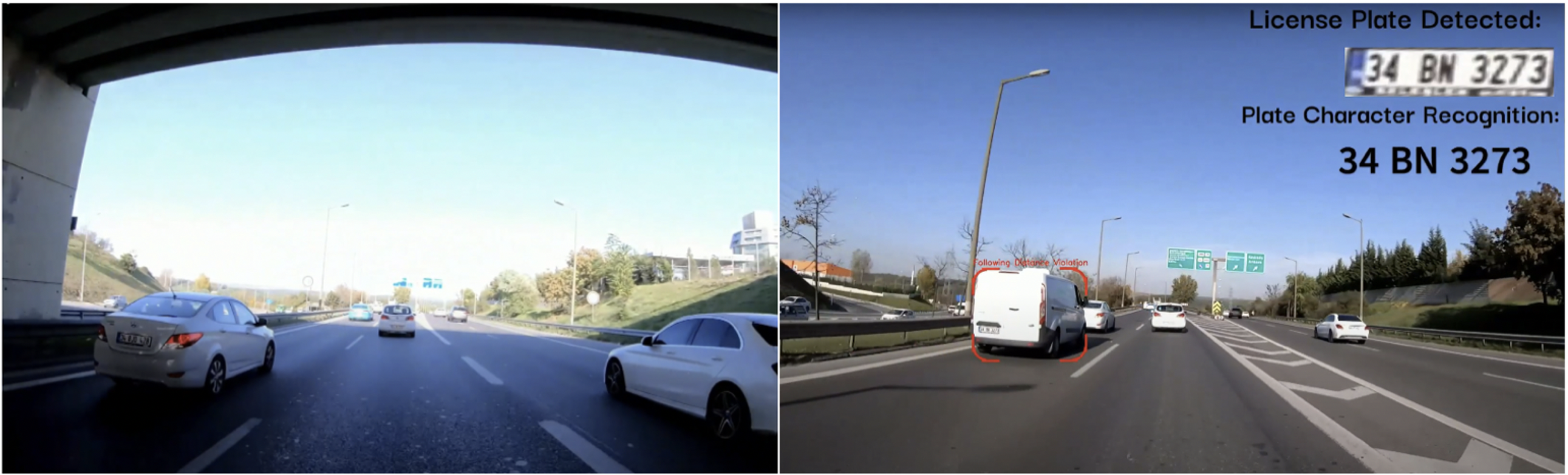}
	\caption{Detection of Violation of Following Distance}
	\label{fig:fig1.png}
\end{figure}

\begin{figure}
	\centering
        \includegraphics[width=14cm]{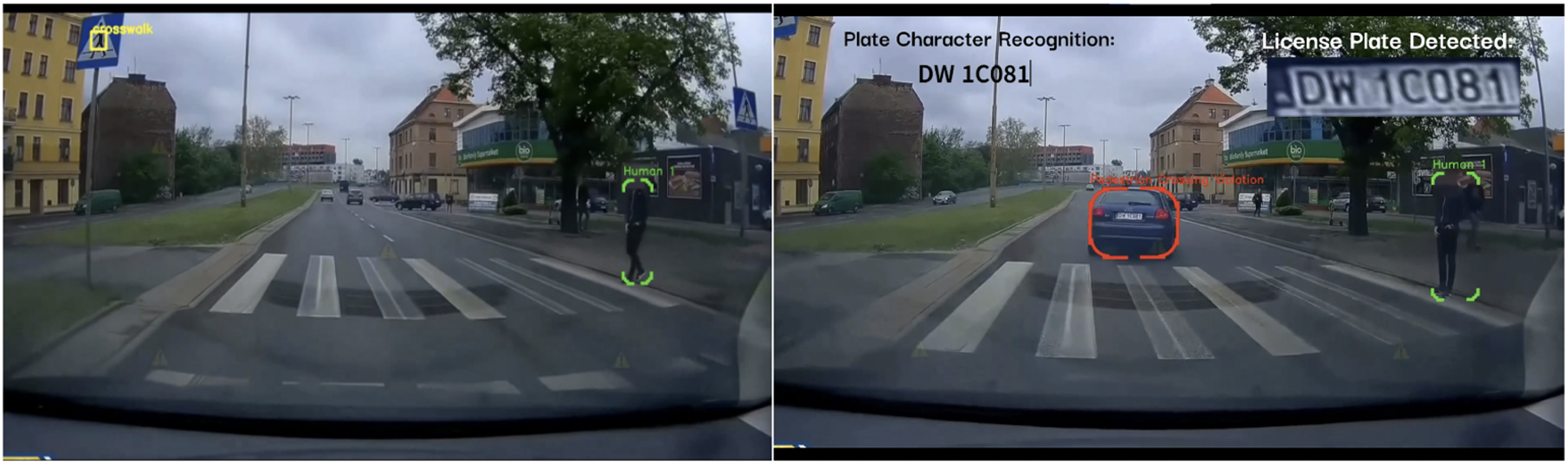}
	\caption{Detection of Pedestrian Crossing Violation}
	\label{fig:fig1.png}
\end{figure}

\begin{figure}
	\centering
        \includegraphics[width=14cm]{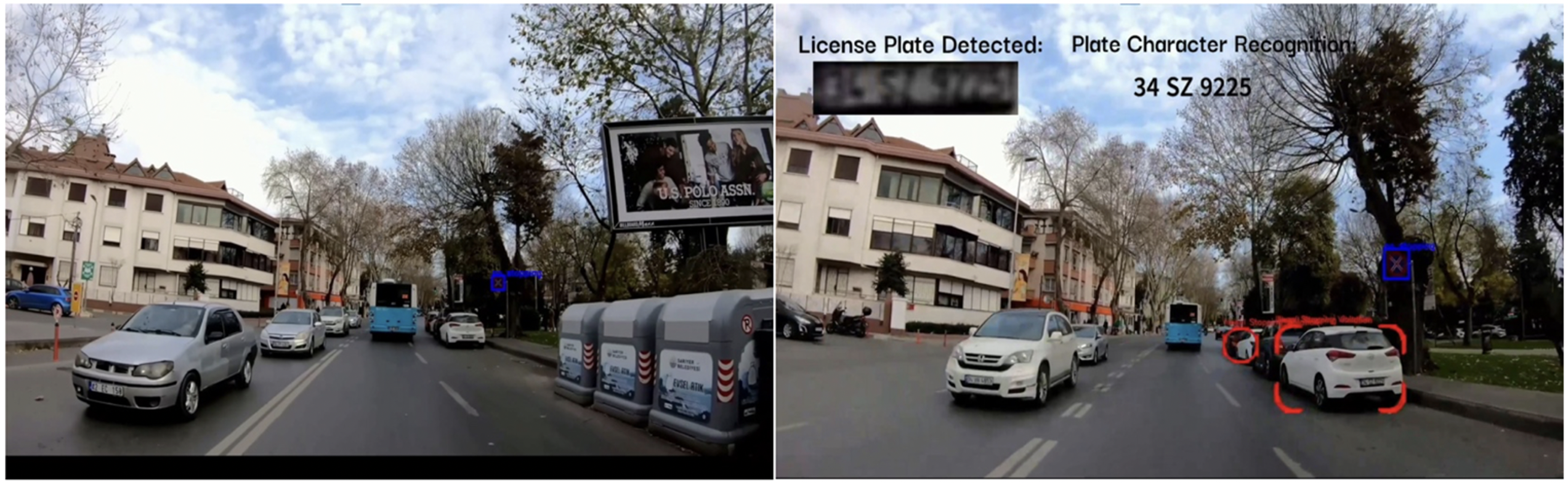}
	\caption{Detection of Illegal Parking Violation}
	\label{fig:fig1.png}
\end{figure}

\begin{figure}
	\centering
        \includegraphics[width=14cm]{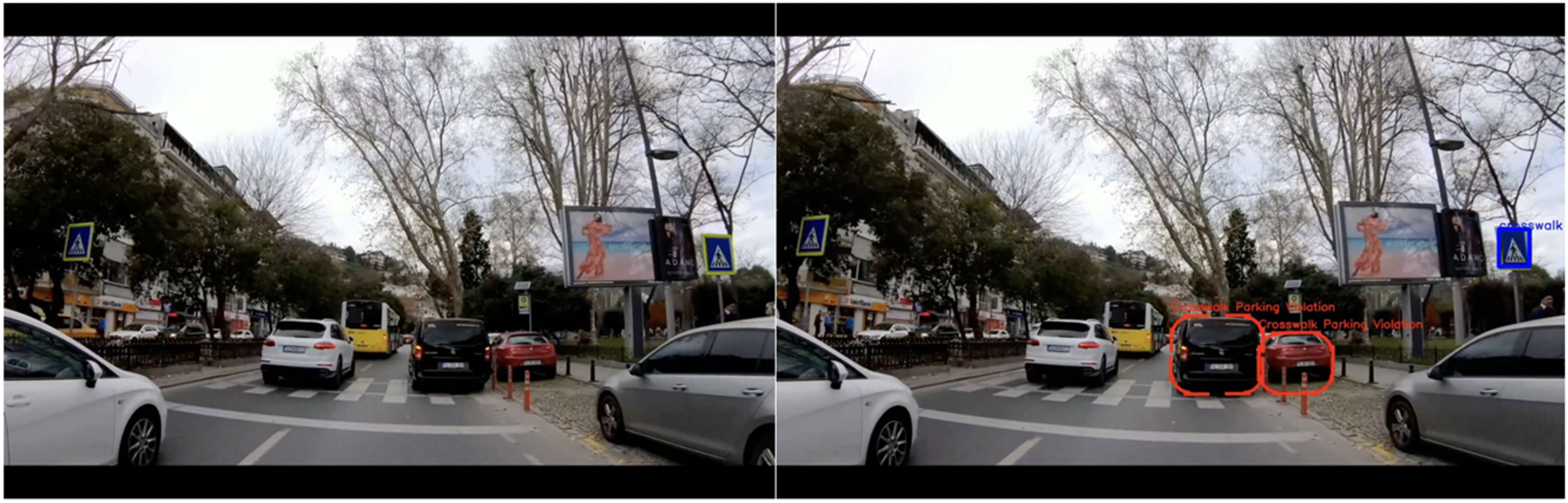}
	\caption{Detection of Marked-Crosswalk Parking Violation}
	\label{fig:fig1.png}
\end{figure}
\end{document}